\documentclass[letterpaper, 10 pt, conference]{ieeeconf}  

\IEEEoverridecommandlockouts                              

\usepackage{cite}
\usepackage{amsmath,amssymb,amsfonts}
\usepackage{algorithmic}
\usepackage{graphicx}
\usepackage{textcomp}
\usepackage{xcolor}

\usepackage{booktabs}
\usepackage{multirow}
\usepackage{hyperref}
\usepackage{amssymb}
\usepackage{bm}
\def\BibTeX{{\rm B\kern-.05em{\sc i\kern-.025em b}\kern-.08em
    T\kern-.1667em\lower.7ex\hbox{E}\kern-.125emX}}

\title{\LARGE \bf NavCMPO: Critic-Guided MeanFlow Policy Optimization for Adaptive Navigation
}

\author{Junjie An$^{1}$, Yi Wu$^{1}$, Xiao Liu$^{1}$, Yiqun Zhou$^{1}$, Yuechen Wu$^{2}$, Xiaoqing Guan$^{1}$, You Wang$^{1,3}$, Guang Li$^{1}$
\thanks{$^{1}$ Zhejiang University, Hangzhou, China. (e-mail: {12332047, 12332084, spanda, 12321119, xiaoqing\_guan, king\_wy, guangli}@zju.edu.cn).}
\thanks{$^{2}$ Shandong University. (e-mail:  wuyuechen@mail.sdu.edu.cn).}
\thanks{$^{3}$ Corresponding author: You Wang.}
}

\begin{document}

\maketitle
\thispagestyle{empty}
\pagestyle{empty}

\begin{abstract}
End-to-end diffusion-based policies have demonstrated strong performance in mapless visual navigation, but their iterative denoising process introduces substantial inference latency, while behavior cloning limits performance to the quality of expert demonstrations. We present NavCMPO, a two-stage adaptive navigation framework that combines few-step MeanFlow trajectory generation, critic-guided refinement, and reinforcement learning fine-tuning. During pre-training, an obstacle proximity prediction task encourages the visual representation to capture obstacle-aware spatial information. To compensate for the degradation in obstacle avoidance caused by few-step generation, Critic-Guided Trajectory Refinement (CGTR) uses gradients from a critic trained with obstacle-point-cloud supervision to refine intermediate trajectories. During adaptation, the MeanFlow policy is fine-tuned using Proximal Policy Optimization with behavior-cloning regularization, while the critic is updated to accommodate embodiment-specific observation changes. Under a matched training budget on the InternVLA-N1 benchmark, NavCMPO achieves an average success rate of 74.7\%, exceeding the retrained NavDP baseline by 6.4 percentage points, while reducing inference latency from 85\,ms to 60\,ms. Experiments on a Unitree Go2 further demonstrate effective sim-to-real transfer.
\end{abstract}


\section{Introduction}

\begin{figure*}[t!]
\centering
\includegraphics[trim={0cm 0.5cm 0cm 0.2cm},clip,width=\textwidth]{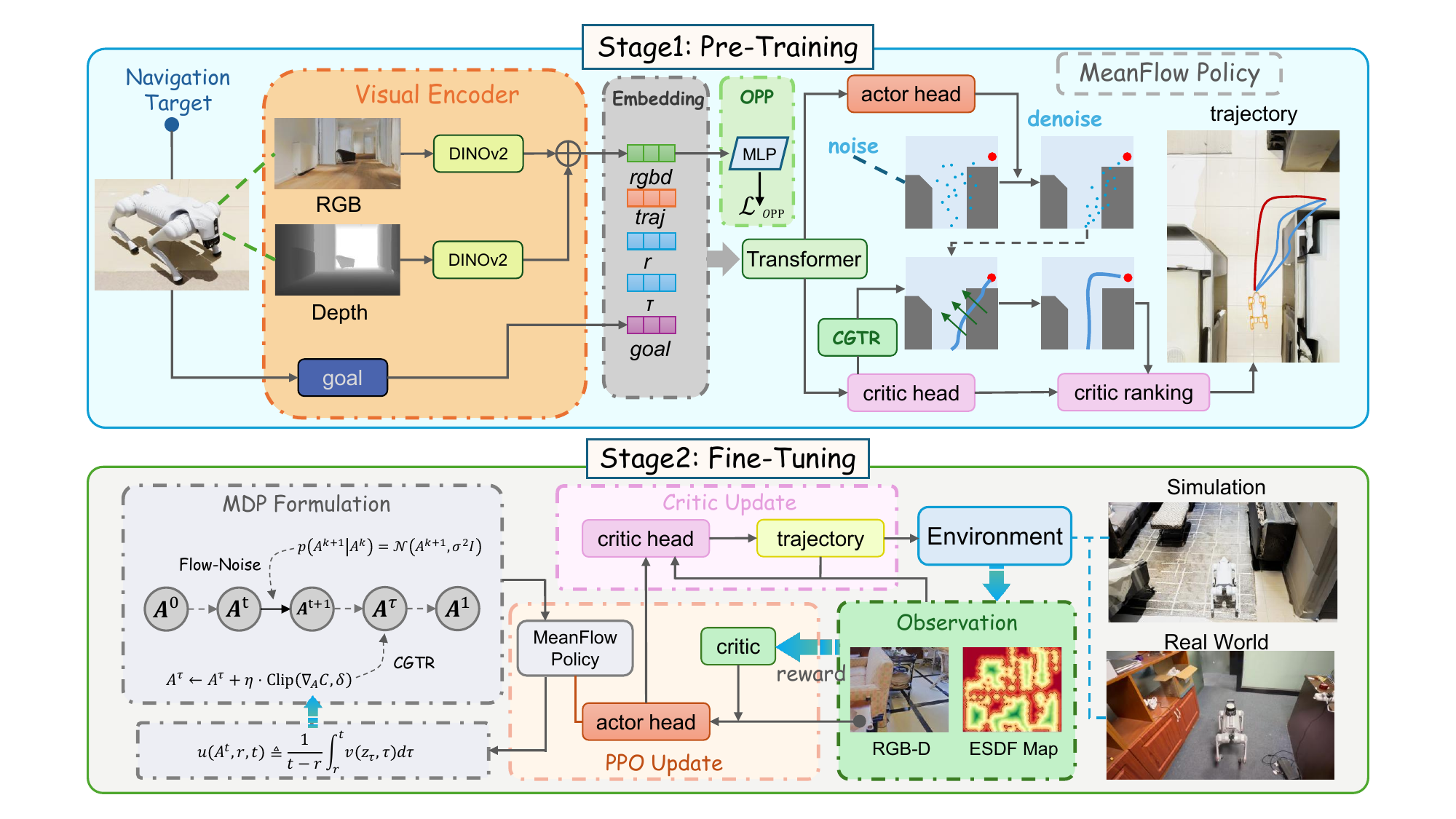}
\caption{Overview of the two-stage NavCMPO framework. In Stage 1, RGB-D observations and the navigation goal are encoded into a conditional representation. The MeanFlow actor generates candidate trajectories, the critic ranks them according to predicted safety and goal progress, and CGTR uses the critic gradient to refine the intermediate flow state. The OPP auxiliary task encourages obstacle-aware visual representations and is removed during inference. In Stage 2, the MeanFlow actor is fine-tuned using PPO with environment rewards and behavior-cloning regularization, while the critic is adapted using geometric supervision from the simulation environment.}
\label{fig:frame}
\end{figure*}

Autonomous visual navigation in cluttered, unstructured environments is a fundamental capability for mobile robots. Classical navigation systems typically rely on modular pipelines consisting of mapping, localization, and path planning, which require significant engineering effort and often fail to adapt to diverse environments and robot platforms\cite{prorobotics}. End-to-end learning-based approaches offer an attractive alternative by directly mapping sensor observations to navigation actions or trajectories, thereby bypassing explicit map construction and reducing the sim-to-real gap \cite{map_vis,Embodnav}.

Current end-to-end visual navigation approaches fall into two paradigms. Reinforcement learning methods \cite{rlnav1,rlnav2,rlnav3} learn navigation policies through trial-and-error interaction, but vision-to-action RL suffers from low sample efficiency, as the network must jointly learn visual perception and motion planning. Behavior cloning offers a more practical alternative, with diffusion-based policies achieving state-of-the-art results. Diffusion Policy~\cite{dp} introduced DDPMs for visuomotor control, and NavDP~\cite{navdp} further adapted this framework for mapless visual navigation with sim-to-real transfer.     

However, diffusion-based navigation policies face three key limitations. First, multi-step denoising incurs high inference latency. NavDP requires 10 iterative denoising steps per planning cycle, each involving a full Transformer decoder forward pass. This accumulated latency hinders real-time deployment, especially on resource-constrained platforms. While recent advances in flow matching and MeanFlow \cite{meanflow,mp1} enable generation with one step, we observe that reducing the number of denoising steps significantly degrades trajectory quality in navigation tasks—the generated trajectories tend to go straight toward the goal direction while ignoring obstacles, as fewer steps provide insufficient opportunities for the model to reason about spatial structure through cross-attention. Second, behavior cloning imposes a performance ceiling. Policies trained purely from expert demonstrations cannot surpass the demonstration quality and are brittle to distributional shifts at deployment. Recent works on RL fine-tuning of generative policies \cite{dppo,dmpo,pirl} have demonstrated that policy gradient methods such as PPO can push diffusion and flow-based policies beyond the imitation learning ceiling, making RL fine-tuning an increasingly prominent paradigm for closing the sim-to-real gap.

In this paper, we present NavCMPO (Critic-Guided MeanFlow Policy Optimization), a two-stage framework for low-latency and adaptive visual navigation. Our focus is not to introduce a new flow-matching or reinforcement-learning algorithm, but to adapt these techniques to visual navigation, where aggressive reduction of generation steps can substantially degrade obstacle avoidance. During pre-training, we combine few-step MeanFlow trajectory generation~\cite{meanflow} with obstacle-aware representation learning and critic-guided trajectory refinement. During adaptation, we fine-tune the policy using PPO following prior generative-policy fine-tuning methods~\cite{dppo,dmpo}, while updating the critic to accommodate embodiment-specific observation changes. This integration improves the quality-latency tradeoff and enables efficient specialization to a target robot without training a visual navigation policy from scratch.

  The main contributions of our proposed framework are threefold:

(1) We adapt MeanFlow-based few-step generation to mapless visual navigation and analyze its quality-latency tradeoff. The resulting five-step policy reduces inference latency from 85\,ms to 60\,ms relative to the retrained NavDP baseline.

(2) We introduce navigation-specific obstacle-aware learning and refinement mechanisms. Obstacle Proximity Prediction improves the spatial representation learned by the visual encoder, while Critic-Guided Trajectory Refinement uses gradients from a learned critic to correct unsafe intermediate trajectories without requiring explicit online trajectory optimization.

(3) We develop a two-stage adaptation scheme that combines PPO fine-tuning, critic adaptation, and behavior-cloning regularization. Under matched training resources, the complete system improves the average benchmark success rate by 6.4 percentage points over NavDP and demonstrates effective transfer to a Unitree Go2.

\section{Related Works}

\subsection{Generative Motion Planning}

Diffusion-based planners model multimodal trajectories but require iterative denoising~\cite{diffuser,dp,dp2,dp3}. Diffuser also shows that gradient-guided sampling supports flexible trajectory planning~\cite{diffuser}. Flow matching constructs straighter generation paths~\cite{flowmatching,flowpolicy}, while Rectified Flow and MeanFlow reduce the required integration steps~\cite{ReFlow,meanflow}. For visual navigation, FlowNav~\cite{flownav} combines flow matching with depth priors, while NaviDiffusor~\cite{navidiffusor} uses explicit TSDF-based cost guidance. In contrast, CGTR refines intermediate trajectories using gradients from a learned critic without constructing an explicit geometric cost at inference.

\subsection{End-to-End Visual Navigation}
iPlanner and ViPlanner integrate differentiable trajectory optimization into learned navigation systems~\cite{iplanner,viplanner}. NoMaD uses goal-masked diffusion for general navigation~\cite{nomad}, while NavDP employs privileged-information guidance for sim-to-real visual navigation~\cite{navdp}. Our work follows the NavDP evaluation setting but focuses on reducing generation latency and adapting the policy to a target embodiment.

\subsection{RL Fine-tuning for Generative Policies}

DPPO applies PPO to diffusion policies~\cite{dppo}, and subsequent methods extend online RL fine-tuning to flow-based policies~\cite{reinflow,flowgrpo,pirl}. DMPO combines MeanFlow with reinforcement learning for efficient manipulation~\cite{dmpo}. In contrast, we study few-step MeanFlow fine-tuning for RGB-D navigation, together with critic-guided refinement, critic adaptation, and behavior-cloning regularization.

\section{Method}
In this section, we propose an efficient and generalizable adaptive navigation framework serving as a local planner for robot navigation. Our method utilizes RGB and depth cameras as inputs and navigation trajectory as output. The overall framework consists of three components: (1) a multimodal Transformer architecture for state encoding; (2) a pre-training phase using MeanFlow Policy for stable few-step generation; (3) an online fine-tuning phase using PPO with behavior cloning regularization. The overall framework is shown in the Fig. \ref{fig:frame}.

\subsection{Problem Formulation}
We formulate the visual navigation task as a Partially Observable Markov Decision Process (POMDP), defined by the tuple $\mathcal{M} = (\mathcal{S}, \mathcal{O}, \mathcal{A}, \mathcal{T}, \mathcal{R}, \gamma)$. At each time step $t$, the robot receives an observation $o_t \in \mathcal{O}$, which comprises a sequence of historical RGB images $I_{t-N:t}$, the current depth image $D_t$, and the navigation goal $g_t$. The policy $\pi_\theta(a_t|o_t)$ generates an action $a_t \in \mathcal{A}$, representing a sequence of future waypoints (trajectory) $\tau_t$. Upon executing the action, the robot transitions to a new state governed by the dynamics $\mathcal{T}$ and receives a reward $r_t$ from the environment. As we aim to enable real-time visual local planning, our objective is to learn an optimal policy $\pi^*_\theta$ that maximizes the expected cumulative reward while maintaining ultra-low inference latency.

\subsection{Meanflow Policy Navigation Pre-Training}
To address the aforementioned challenges, visual navigation involves an extremely large state space (high-dimensional images) and a continuous action space\cite{acrl,rlks}. Training reinforcement learning (RL) agents from scratch requires millions of random exploration steps within the environment to acquire basic behaviors such as obstacle avoidance or target approaching. This process is computationally expensive and time-consuming. Conversely, imitating expert trajectory planning policies offers significant advantages; by learning from large-scale datasets, robots can acquire implicit expert preferences, such as generating smooth trajectories and avoiding obstacles.

In the pre-training phase, we utilize a meanflow policy as the local trajectory planning strategy. We adopt a unified transformer \cite{navdp} architecture as the policy backbone to process high-dimensional visual inputs.

\textbf{Multi-Modal Encoder}: To capture spatial geometry and temporal context, we employ two DINOv2 ViT-S encoders~\cite{dinov2} for RGB and depth observations, respectively. The historical RGB images are processed by the RGB encoder, while the current depth image is replicated into three channels and processed by the depth encoder. The resulting RGB and depth tokens are fused through a lightweight Transformer decoder and projected into the conditional representation $h_{cond}=\phi_\theta(o_t)$.

\textbf{Obstacle Proximity Prediction(OPP)}: 
To enable the visual encoder to learn spatially-aware feature representations, we introduce the OPP auxiliary task. OPP partitions the space surrounding the robot into 12 uniform angular sectors and predicts the distance to the nearest obstacle within each sector. OPP auxiliary task employs an MLP to project $h_{cond}$ into the predicted distance vector $d_{pred}$.

Ground truth labels are derived directly from the obstacle point clouds in the dataset. We compute the relative distance and azimuth of each obstacle point. These points are then assigned to their corresponding sectors based on angle, and the minimum distance within each sector $d_{min}$ is selected as the label. The OPP loss is defined as:
\begin{equation}
    \mathcal{L}_{OPP} = \text{SmoothL1}({d}_{pred},{d}_{min})
\end{equation}

\textbf{Policy Head}: Our policy head predicts the velocity field $u_\theta(z_t, t, h_{cond})$ within the flow matching process\cite{meanflow}, where $t \in [0, 1]$ represents the flow time and $z_\tau$ denotes the flow state. This improvement enables a transition from multi-step denoising to efficient flow-based generation.

We model trajectory generation as a deterministic flow from a standard Gaussian noise distribution $\epsilon \sim \mathcal{N}(0, I)$ to the expert trajectory distribution $a_{tgt}$. We define the linearly interpolated path as $z_t = (1-t)a_{tgt} + t \epsilon$. The training objective is to train the network $u_\theta$ to predict the average velocity field capable of transporting noise to data in a single step ($K=1$). The average velocity field satisfies the following equation:

\begin{equation}
    \frac{d}{dt}u(z_t, r, t) = v(z_t, t)\partial_z u + \partial_t u
\end{equation}

The MeanFlow loss function is defined as:

\begin{equation}
    \mathcal{L}_{MF}(\theta) = \mathbb{E} \left[ \| u_\theta(z_t, r, t, h) - u_{tgt} \|^2_2 \right]
\end{equation}

where $u_{tgt} = (\epsilon - a_{tgt})$ represents the target velocity directed from data to noise.

As shown in Sec.~\ref{exp:one_step}, directly applying one-step MeanFlow generation results in a low navigation success rate, even when dispersion regularization is used~\cite{mp1,dmpo}. In particular, the generated trajectory may point directly toward the goal while failing to represent alternative paths around an obstacle. We therefore use multi-step MeanFlow integration to provide additional opportunities for visual features to interact with the trajectory representation. We divide the generation interval $[0,1]$ into $T$ segments, where $t_k=k/T$, and update the flow state as follows:

\begin{equation}
    z_{t_{k+1}} = z_{t_k} - \frac{1}{T} \cdot u_\theta(z_{t_k}, t_k, t_{k+1}, h_{cond})
\end{equation} 

\textbf{Critic Head}\label{Critic_Head}: During inference, the policy generates $N$ candidate trajectories in parallel. A critic head network $C(z_t, h_{cond})$ evaluates each trajectory and predicts a quality score\cite{navdp}. The critic shares the same Transformer decoder as the policy head, enabling it to jointly assess trajectory safety and goal-directed progress. The scoring function is derived from obstacle point cloud supervision:

\begin{equation}\label{eq:critic}
    s(\tau) = -\lambda_c \cdot \frac{1}{N_\tau}\sum_{i}^{N_\tau} \left[D(\tau_i) < d_{safe}\right] + \lambda_g \cdot \Delta d_{goal}
\end{equation}

where $D(\tau_i)$ denotes the minimum distance between waypoint $\tau_i$ and the observed obstacle points, $d_{safe}$ is the collision-safety threshold, and $\Delta d_{goal}$ measures the goal progress from the first waypoint to the last waypoint. The indicator term penalizes waypoints that violate the safety threshold.

We set $\lambda_c=5.0, \lambda_g=0.1$ to prioritize obstacle avoidance, while the goal term serves as a tiebreaker among equally safe trajectories. The critic is trained via mean squared error between predicted and ground-truth scores:
\begin{equation}\label{eq:critic_loss}
    \mathcal{L}_{C} = \|C(\tau_{exp}, h) - s(\tau_{exp})\|^2 + \|C(\tau_{aug}, h) - s(\tau_{aug})\|^2
\end{equation}
where $\tau_{exp}$ and $\tau_{aug}$ denote the expert trajectory and its randomly perturbed counterpart, respectively.

The total pre-training loss is defined as :
\begin{equation}
\begin{split}
    \mathcal{L}_{Stage1} = \lambda_{MF}\mathcal{L}_{MF} + \lambda_{OPP}\mathcal{L}_{OPP} + \lambda_{Critic}\mathcal{L}_{C}\\
    + \lambda_{aux}\mathcal{L}_{aux}
\end{split}
\end{equation}

where $\mathcal{L}_{aux}$ represents the positional encoding loss \cite{navdp}. The MeanFlow denoising network is updated via backpropagation using the Jacobian-Vector Product (JVP)\cite{meanflow}. This formulation ensures that the policy effectively captures diverse navigation modes.

\textbf{Critic-Guided Trajectory Refinement(CGTR)}:
Although the MeanFlow Policy can improve the success rate under few-step denoising, the limited number of generation steps still poses potential safety risks. To address this, we propose a trajectory correction method based on Critic gradients. In the penultimate step of the denoising process, the gradient signals from a pre-trained Critic network are utilized to explicitly refine the intermediate trajectory.   
\begin{equation}
    z_\tau \leftarrow z_\tau + \eta \cdot \text{Clip}\left(\nabla_{z} C(z_t, h_{cond}), \delta\right)
    \label{eq:CGTR}
\end{equation}

where $C(\cdot)$ is the critic trained with obstacle-point-cloud supervision, $\eta$ is the guidance coefficient, and $\delta$ is the gradient-clipping threshold. We apply CGTR at the penultimate generation step, denoted by $\tau$. Because a larger critic value represents a safer and more goal-directed trajectory, the positive critic gradient provides a local direction for improving the intermediate trajectory. This mechanism is inspired by classifier guidance~\cite{claguide}, but uses a navigation critic rather than a class-conditional image classifier.

\subsection{Online RL Fine-tuning for Meanflow Policy}

In Stage 2, we optimize the robot navigation policy using reinforcement learning in simulation. We formalize the fine-tuning process as the optimization of MeanFlow Policy. To facilitate exploration, we inject stochastic noise into the last few flow process during training, parameterizing the action distribution as a Gaussian centered on the MeanFlow prediction\cite{dppo,fpo,pirl,dmpo}:

\begin{equation}
    p(z_{t_{k+1}}|z_{t_k}) = \mathcal{N}\left(z_{t_k}-\frac{1}{T} u(z_{t_k},t_{k+1},t_k,h), \sigma^2 I\right)
\end{equation}

We employ Proximal Policy Optimization (PPO) to update the policy parameters $\theta$. The denoising log-probability is defined as:
\begin{equation}
    \log \pi(z_{t_{k+1}} | z_{t_k}, h) = \log \mathcal{N}\left(z_{t_{k+1}}|\boldsymbol{\mu}_k, \sigma^2 \mathbf{I}\right) ,\ t_{k}\neq \tau
\end{equation}
With the incorporation of CGTR at denoising step $\tau$, the log-probability becomes:
\begin{equation}
    \log \pi(z_{\tau+1} | z_{\tau}, \mathbf{s}) = \log \mathcal{N}\left(z_{{\tau+1}}+\mathcal{C}|\boldsymbol{\mu}_k, \sigma^2 \mathbf{I}\right)
\end{equation}
where $\mathcal{C}$ is the CGTR correction term (\ref{eq:CGTR}).

\textbf{Reward}: For the robot navigation task, we design a dense reward function defined as:

\begin{equation}
    R_t = \beta_{p}R_{progress} + \beta_{s}R_{safety} + \beta_{g}R_{goal}\label{eq:reward}
\end{equation}

where $R_{progress} = d_{goal}^{t-1} - d_{goal}^t$ measures the reduction in Euclidean distance to the goal, providing dense feedback at each step. $r_{safety}$ leverages the Euclidean Signed Distance Field (ESDF) map to penalize collisions and unsafe proximity:
\begin{equation}
    R_{safety} = -1, \ \text{if} \ D(p_t)<d_{safe}
\end{equation}
where $D(p_t)$ is the ESDF map values of the robot at the current position. $R_{goal} = 1$ is a bonus awarded upon reaching within 0.5m of the goal.

Following PPO-based generative-policy fine-tuning~\cite{dppo,dmpo}, we optimize the standard clipped surrogate objective:
\begin{equation}
\mathcal{L}_{PPO}
=-\mathbb{E}\left[\min\left(\rho_{t,k}\hat{A}_t,
\operatorname{clip}(\rho_{t,k},1-\epsilon,1+\epsilon)\hat{A}_t\right)\right],
\end{equation}
where $\rho_{t,k}=\pi_\theta(z_{t_{k+1}}\mid z_{t_k},h)/\pi_{\theta_{\mathrm{roll}}}(z_{t_{k+1}}\mid z_{t_k},h)$, $\hat{A}_t$ is the estimated advantage, and $\epsilon$ is the clipping coefficient. Here $\pi_{\theta_{\mathrm{roll}}}$ is the rollout policy, while $\pi_{ref}$ below denotes the frozen BC reference policy.

\textbf{Critic Update}: Changes in camera placement, field of view, and robot embodiment can shift the visual observations encountered during deployment, thereby reducing the accuracy of the pre-trained critic. Because the critic is used both to rank candidate trajectories and to compute the CGTR direction, we continue to optimize it during Stage 2 using the same score definition and supervised critic loss as in Eqs.~\ref{eq:critic} and~\ref{eq:critic_loss}.

To retain the navigation prior learned during pre-training, we keep a frozen reference policy $\pi_{ref}$ and penalize deviations of the fine-tuned policy from its generated trajectories:

\begin{equation}
\mathcal{L}_{BC}
=
\mathbb{E}\left[
\|\tau_{\pi_{ref}}-\tau_{\pi_\theta}\|_2^2
\right].
\end{equation}

PPO clipping and BC regularization serve different purposes. PPO clipping constrains each update relative to the policy that collected the current rollout, whereas $\mathcal{L}_{BC}$ anchors the policy to the fixed pre-trained navigation prior throughout fine-tuning. The latter reduces long-term drift and catastrophic forgetting over multiple PPO updates.

The final optimization objective for the second stage is defined as:

\begin{equation}
    \mathcal{L}_{Stage2} = \mathcal{L}_{PPO} + \lambda_{BC} \mathcal{L}_{BC} + \lambda_{Critic2} \mathcal{L}_{C}
\end{equation}

This formulation enables the robot to adapt to complex dynamics and optimize trajectory smoothness while retaining the robust visual representations learned from large-scale datasets.

\section{Experiments}

\begin{table*}[t]
\centering
\caption{Point-goal navigation results under the reported evaluation protocols. Best results are shown in \textbf{bold}; ``--'' indicates that the corresponding result is not available or was not reported.}
\label{tab:main}
\begin{tabular}{lcccccc}
\toprule
\multirow{2}{*}{Method} & Cluttered-Hard & Home & Commercial & Average & Inference \\
  & SR / SPL & SR / SPL & SR / SPL & SR / SPL & (ms) \\
\midrule
DD-PPO & - & - & - & 8.6 / 8.5 & - \\
iplanner & - & - & - & 54.1 / 51.2 & - \\
Viplanner & 68.4 / 56.8 & 53.7 / 51.2 & 68.1 / 66.0 & 62.4 / 58.2 & 30 \\
NavDP & 72.8 / 51.2 & 60.3 / 54.7 & 74.1 / 70.5 & 68.3 / 60.3 & 85 \\
MP1 & 77.6 / \textbf{61.4} & 46.4 / 43.2 & 62.5 / 59.2 & 59.1 / 53.2 & \textbf{30} \\
NavCMP & 76.8 / 59.2  & 61.1 / 55.8 & 73.4 / 70.1 & 69.2 / 62.2 &  60 \\
\textbf{NavCMPO} & \textbf{82.8} / 59.4 & \textbf{66.9 / 60.6} & \textbf{78.4 / 72.3} & \textbf{74.7 / 65.1} & 60 \\
\bottomrule
\end{tabular}
\end{table*}

To compare our method with NavDP, we adopt NavDP's evaluation benchmark and assess our algorithm's performance in both simulation and real-world settings. The simulation test environment is built on IsaacSim, with Dingo wheeled robot. The test environments include Cluttered-Easy, Cluttered-Hard, Home, and Commercial scenes. Our primary evaluation focuses on point-to-point navigation tasks. Additionally, we conduct real-world experiments using a Unitree Go2 quadruped robot across  three scenarios: indoor static scenes, indoor dynamic scenes, and outdoor scenes, likewise evaluating point-to-point navigation tasks.

  Evaluation Metrics. Following standard navigation evaluation protocols, we
  report:
\begin{itemize}
    \item \textbf{Success Rate (SR)}: The percentage of episodes in which the robot reaches within 1m of the goal.
    \item \textbf{Success weighted by Path Length (SPL)}: SR weighted by the ratio of optimal path length to actual path length, measuring path efficiency.
    \item \textbf{Inference Time}: The average wall-clock time per planning step (measured on an NVIDIA RTX 4090).
\end{itemize}
  
\subsection{Simulation Results}


\begin{figure}[h]                                                                                              
\centering                             
\includegraphics[trim={0.6cm 0cm 0.6cm 0cm},width=0.66\columnwidth]{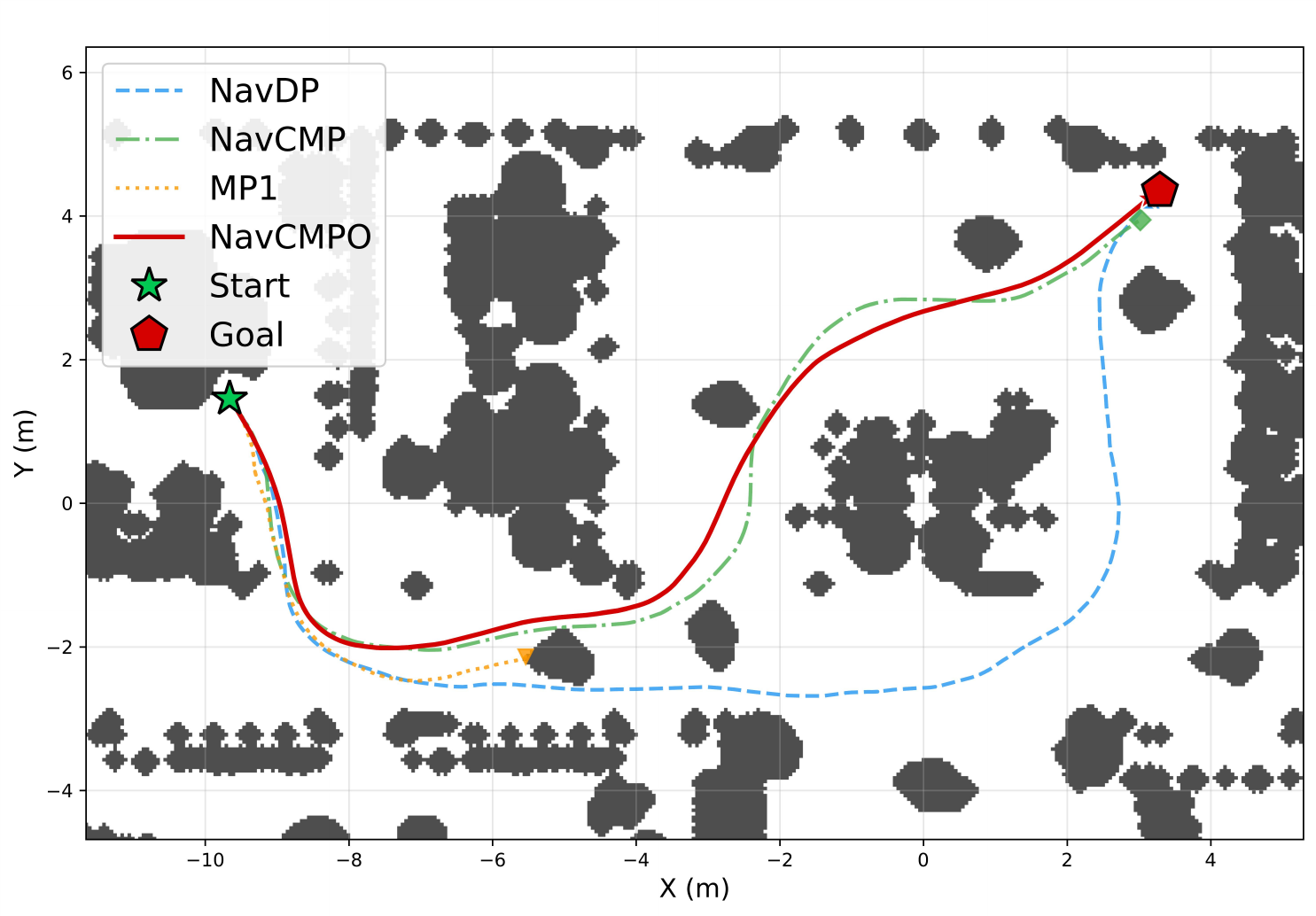}
\hfill                                                                                                         
\includegraphics[width=0.32\columnwidth]{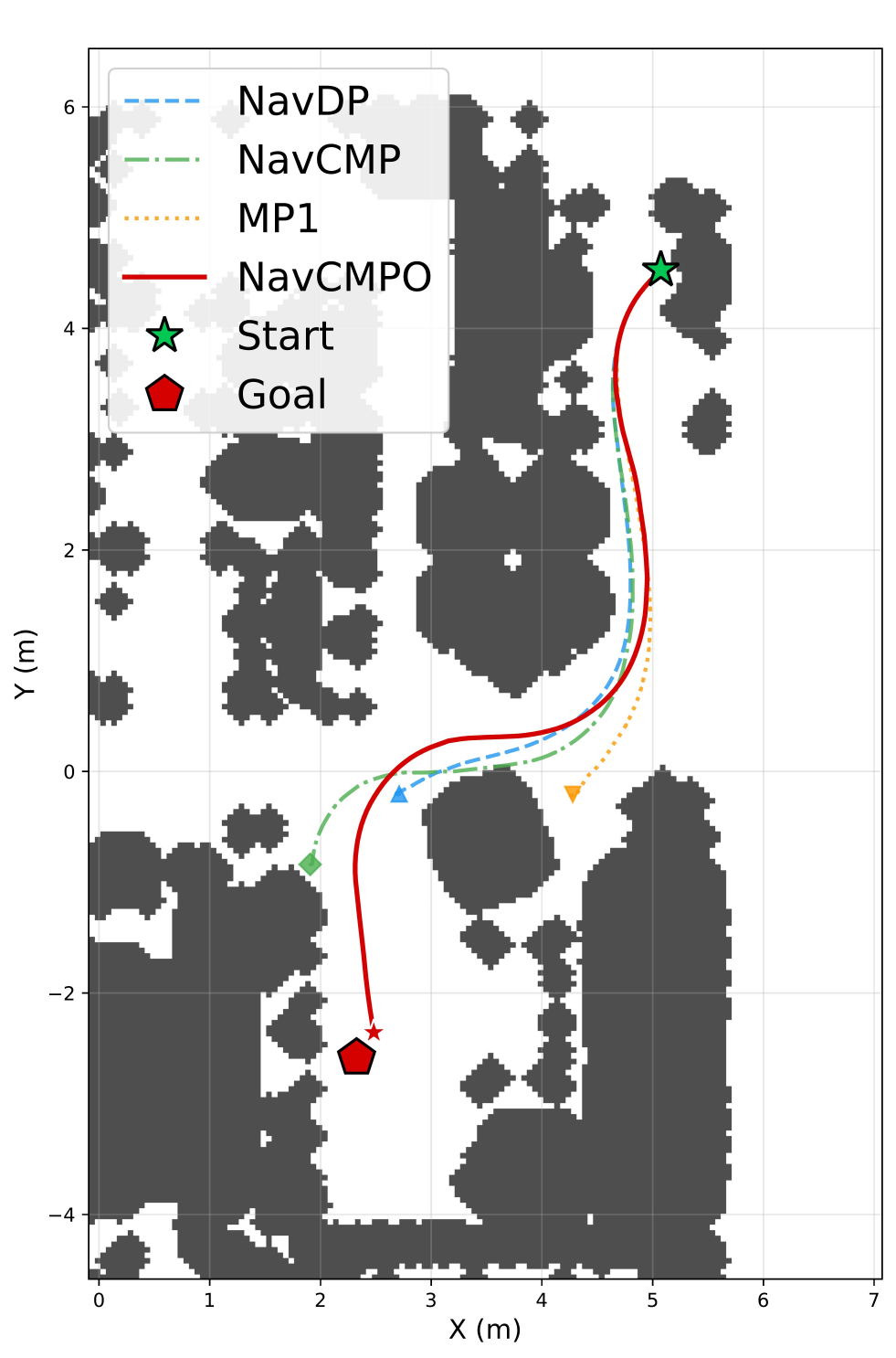} 
\caption{Qualitative trajectory comparison in two representative indoor scenes. Left: Long distance task; Right: Narrow and complex indoor scene with other baselines failing. Our method (NavCMPO, red) generates safer paths while successfully avoiding obstacles.}
\label{fig:traj_comparison}
\end{figure}

In this section, we conducted a series of comparative and ablation studies in simulation. To ensure a fair comparison, all algorithms were trained using identical computational resources and training durations. The training settings are: Pre-training: 8$\times$A800 GPUs, batch size of 128, trained for 24 hours;
Fine-tuning: A single RTX 4090 GPU, batch size of 8, trained for 12 hours.
Thus, Table~\ref{tab:main} reports a matched-compute comparison with our retrained NavDP baseline and does not imply superiority over the original NavDP across its larger-scale training or broader embodiment coverage~\cite{navdp}. We fine-tune the test robot using an indoor environment from InternUtopia. 
The test benchmarks in simulation include three types of scenarios: Cluttered Hard, Home, and Commercial, among which Cluttered Hard has 500 episodes, while Home and Commercial each have 1000 episodes.

In the comparative experiments, our baselines include:

\begin{itemize}
    \item \textbf{DD-PPO}~\cite{ddppo}: A reinforcement-learning navigation baseline trained using decentralized distributed PPO.
    \item \textbf{iPlanner}~\cite{iplanner}: A differentiable local-planning method that learns to generate collision-free trajectories.
    \item \textbf{Viplanner}: Visual semantic instruction planner.
    \item \textbf{NavDP}: Uses Diffusion Policy for behavior cloning navigation with 10 denoising steps.
    \item \textbf{MP1}: Uses One Step Dispersive MeanFlow Policy\cite{mp1} for behavior cloning navigation.\label{exp:one_step}
    \item \textbf{NavCMP(ours)}: Uses Critic-Guided MeanFlow Policy for behavior cloning navigation with 5 denoising steps.
\end{itemize}

The experimental results are presented in Table \ref{tab:main}. MP1 (one-step MeanFlow) achieves the fastest inference (30ms) but suffers from significantly lower success rates in complex environments (Home: 46.4\%, Commercial: 62.5\%), confirming that single-step generation is insufficient for navigation tasks requiring spatial reasoning. Our NavCMP (5-step Critic-Guided MeanFlow) recovers performance to match NavDP across all scenes while reducing inference time to 60ms. Flow-based models tend to generate more direct trajectories, as illustrated in Fig. \ref{fig:traj_comparison}, whereas diffusion models often produce more tortuous paths due to injected noise. The integration of reinforcement learning proves crucial: NavCMPO achieves the highest average SR and SPL, outperforming NavDP by 6.4\% SR and 4.8\% SPL.

\begin{figure}[h]                                                                                              
\centering                             
\includegraphics[width=0.48\columnwidth]{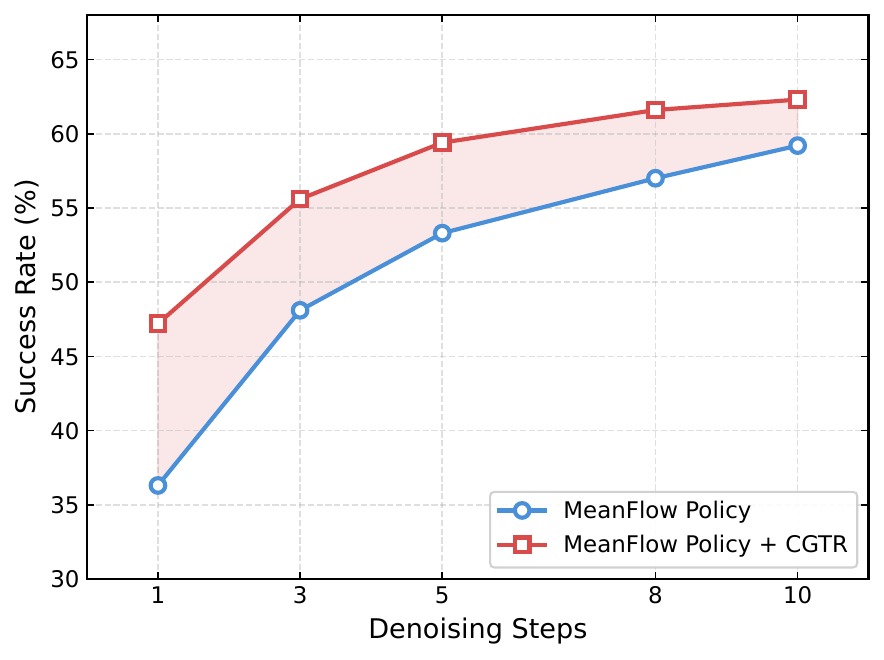}
\hfill                                                                                                         
\includegraphics[width=0.48\columnwidth]{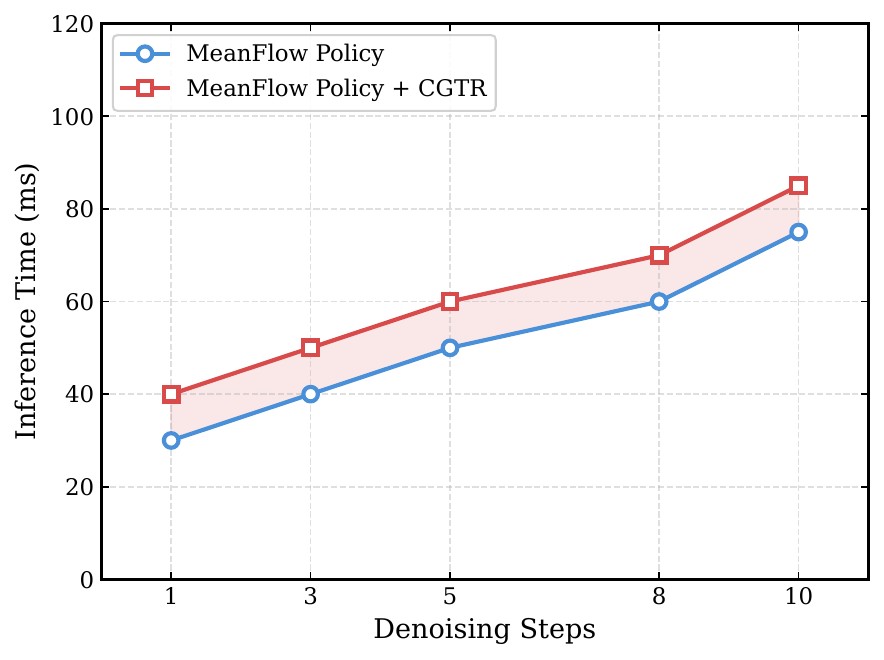} 
\caption{Left: Success rate versus denoising steps with and without CGTR. Right: Inference time versus denoising steps with and without CGTR.}
\label{fig:denoising_step}
\vspace{-0.5cm}
\end{figure}

Regarding inference latency, NavCMPO (60ms) operates at real-time frequency while outperforming NavDP (85ms) by 6.4\% in average SR. The 5-step MeanFlow formulation provides the best quality-latency tradeoff, as shown in Fig.~\ref{fig:denoising_step}. Overall, our method achieves state-of-the-art performance with an average success rate of 74.7\% and SPL of 65.1\%.

Simultaneously, we conducted an ablation study to analyze the effectiveness of our proposed improvements. The variants in the ablation study include:
\begin{itemize}
    \item \textbf{FlowNav}: Replace the MeanFlow Policy with Flow Matching Policy, with 5 denoising steps.
    \item \textbf{w/o OPP}: The MeanFlow Policy is pre-trained without Obstacle Proximity Prediction.
    \item \textbf{w/o CGTR}: The MeanFlow Policy is pre-trained without Critic-Guided Trajectory Refinement.
    \item \textbf{w/o RFT}: The model is deployed without reinforcement learning fine-tuning.
    \item \textbf{w/o CriUp}: The Critic Head is not updated in stage 2.
    \item \textbf{w/o BC Loss}: The Behavior Cloning (BC) loss is excluded from the total loss function during reinforcement learning training.
\end{itemize}

\begin{table}[h]
\centering
\caption{Ablation Study results.}
\label{tab:ablation}
\begin{tabular}{lccc}
\toprule
\multirow{2}{*}{Method} & Easy & Hard  & Average \\
 & SR / SPL & SR / SPL &  SR / SPL \\
\midrule
FlowNav & 60 / 53.1  & 42 / 40.4 &  51.0 / 46.8  \\
w/o OPP & 71 / 64.6  & 58 / 51.4 &  64.5 / 58.0  \\
w/o CGTR & 67 / 59.2 & 53 / 49.2 &  60.0  / 54.2  \\
w/o RFT & 73 / \textbf{65.9} & 59 / 52.9 & 66.0 / 59.4 \\
w/o CriUp & 79 / 64.3 & 66 / 59.4 & 72.5   / 61.2 \\
w/o BC Loss & \textbf{81} / 58.9 & 66 / 54.7 &  73.5 / 56.8 \\
\textbf{NavCMPO} & 79 / 65.2 & \textbf{68 / 62.1} & \textbf{73.5 / 64.2}  \\
\bottomrule
\end{tabular}
\end{table}

Table~\ref{tab:ablation} reports results on 100 indoor scenes under Easy and Hard settings. FlowNav achieves the lowest average SR (51.0\%), below the one-step MeanFlow policy MP1, indicating that MeanFlow provides better few-step generation quality. Removing CGTR causes the largest SR drop (13.5\%), while removing OPP reduces SR by 9.0\%, confirming the importance of critic-guided refinement and obstacle-aware representation learning. Without RL fine-tuning, SR decreases by 7.5\%, although Easy-scene SPL slightly improves to 65.9\%, suggesting that the behavior-cloned policy produces efficient but less robust paths. Updating the critic improves SR/SPL by 1.0/3.0 percentage points, while removing BC regularization increases Easy SR to 81\% but reduces average SPL by 7.4\%. NavCMPO achieves the best Hard-scene and average performance, demonstrating the complementary effects of its components.

\begin{figure}[h]
\centering
\includegraphics[trim={0cm 5cm 9.5cm 0.0cm},clip,width=\columnwidth]{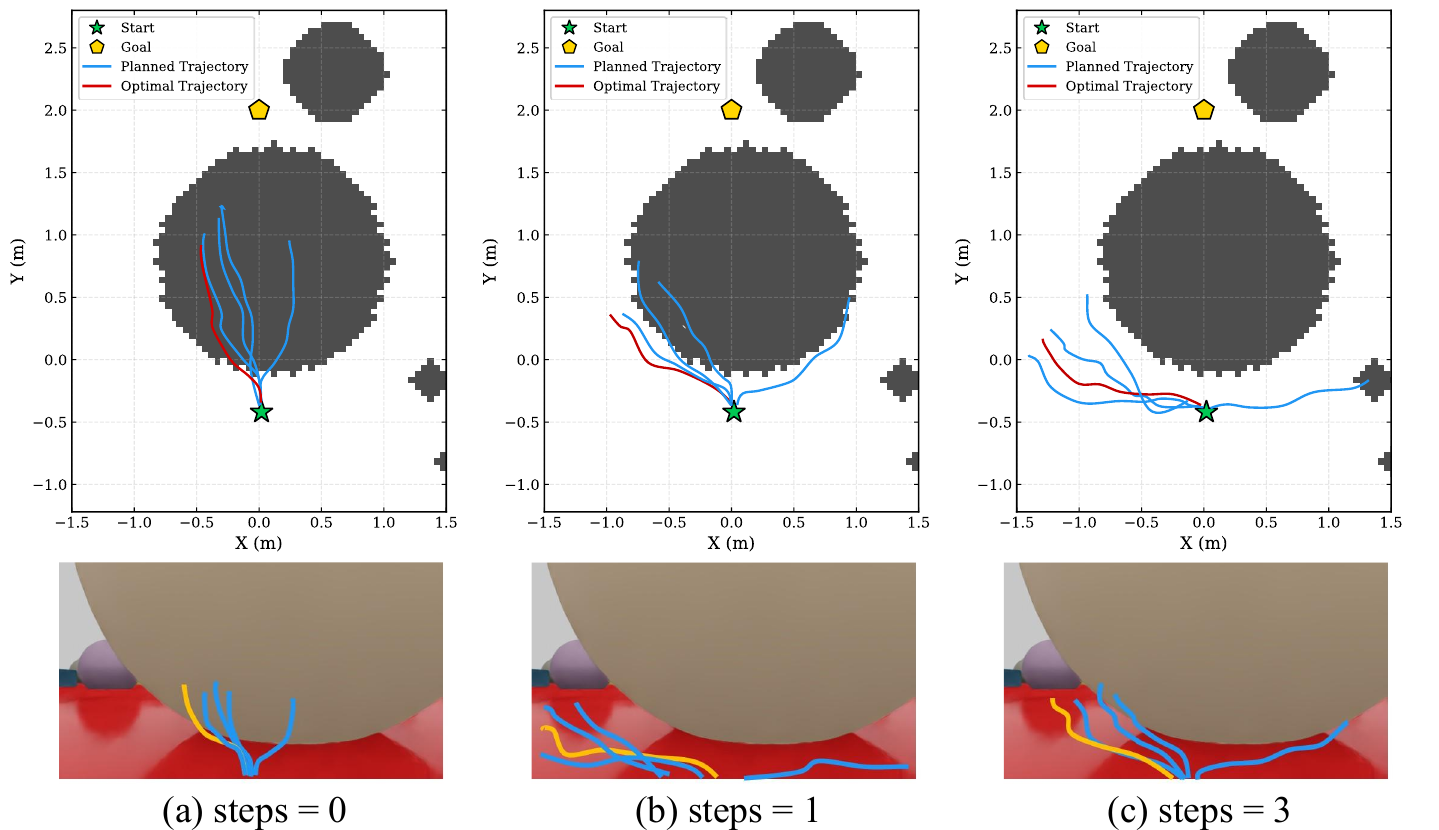}
\caption{Effect of CGTR steps on trajectory quality. The top row shows planned trajectories (blue) and the optimal trajectory (red) on a top-down occupancy map, while the bottom row shows the image in camera. (a) steps = 0: without CGTR, the planned trajectories penetrate the obstacle region. (b) steps = 1: a single CGTR step effectively steers trajectories around the obstacle. (c) steps = 3: excessive guidance causes over-correction.}
\label{fig:CGTR}
\end{figure}

\begin{figure}[h]
\centering
\includegraphics[trim={0cm 0cm 0cm 0.0cm},clip,width=0.6\columnwidth]{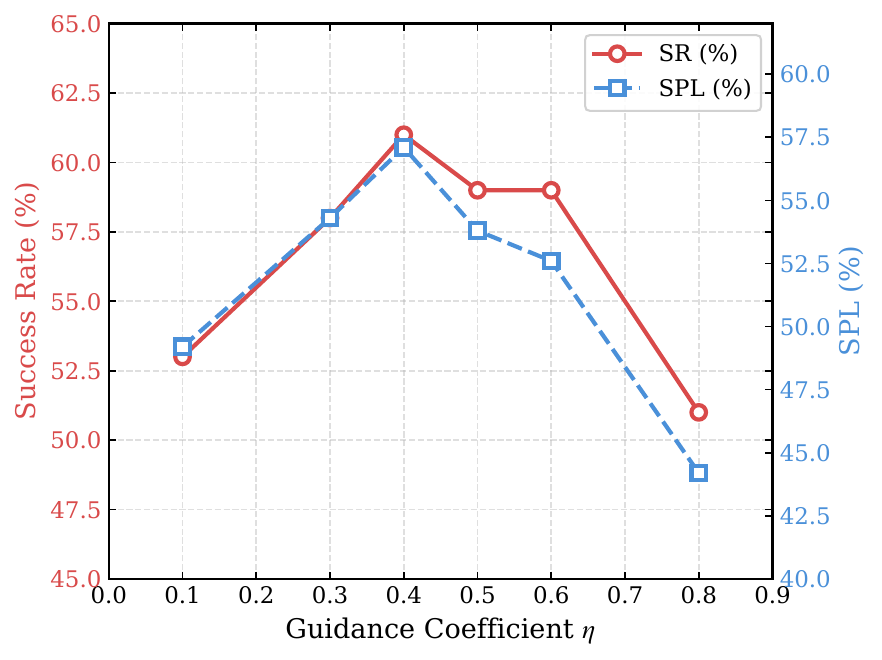}
\caption{Sensitivity analysis of CGTR guidance coefficient $\eta$ with fixed clipping threshold $\delta=1.0$ and step =1. Both SR and SPL peak at $\eta=0.4$.}
\label{fig:cgtr_eta}
\vspace{-0.85cm}
\end{figure}

In the pre-training stage, we conduct ablation experiments to determine the optimal generation and guidance configurations. As shown in Fig.~\ref{fig:denoising_step}, single-step MeanFlow achieves only 36.3\% SR due to trajectory collapse, while 5-step generation reaches 53.3\% SR with significantly lower latency than 10-step (59.2\%). With CGTR enabled, the 5-step configuration achieves 59.4\% SR, matching the 10-step baseline without CGTR, confirming that CGTR effectively compensates for fewer denoising steps at minimal overhead (~10ms). For CGTR configuration, Fig.~\ref{fig:CGTR} demonstrates that single-step guidance optimally steers trajectories around obstacles, whereas multi-step guidance causes excessive correction and trajectory instability. The sensitivity analysis of guidance coefficient $\eta$ (Fig.~\ref{fig:cgtr_eta}) reveals a clear peak at $\eta=0.4$ (61\% SR, 57.1\% SPL), with smaller values providing insufficient correction and larger values causing over-correction ($\eta=0.8$ drops to 51\% SR). Based on these results, we adopt 5 denoising steps with single-step CGTR at $\eta=0.4$ and $\delta=1.0$ as our default configuration.

\begin{figure}[h]
\centering
\includegraphics[trim={0cm 0cm 0cm 0cm},clip,width=0.8\columnwidth]{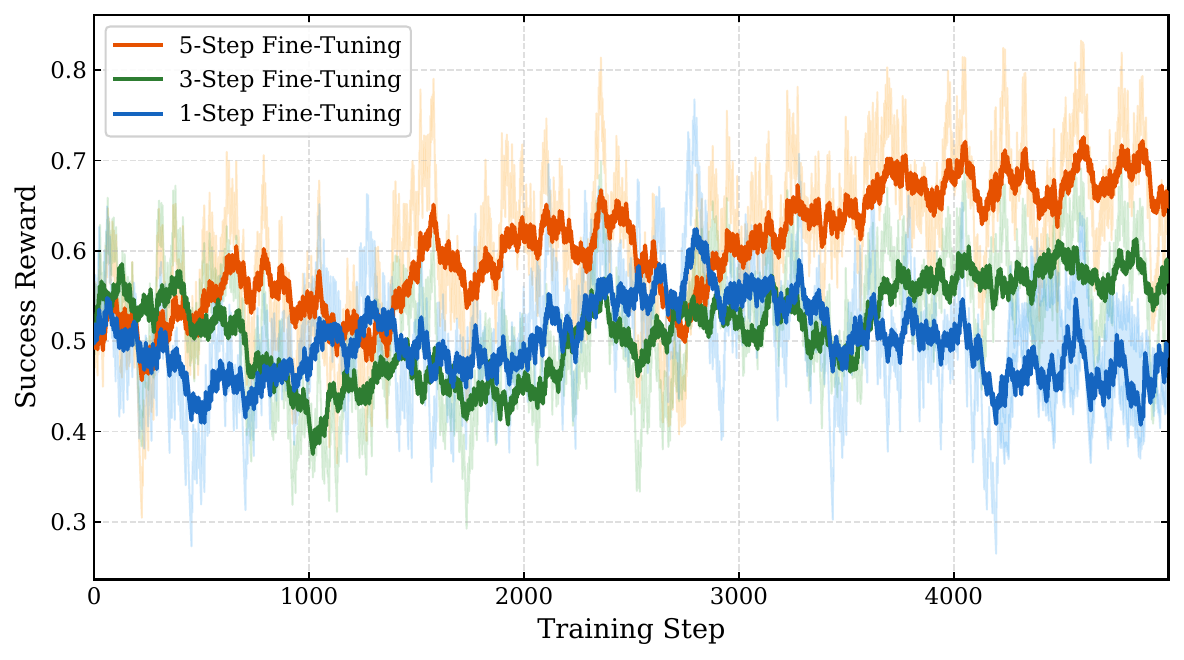}
\caption{PPO fine-tuning reward curves under different numbers of trainable flow integration steps.}
\label{fig:reward}
\vspace{-0.4cm}
\end{figure}

\begin{figure*}[t!]
\centering
\includegraphics[trim={0cm 1cm 3.5cm 5cm},clip,width=\textwidth]{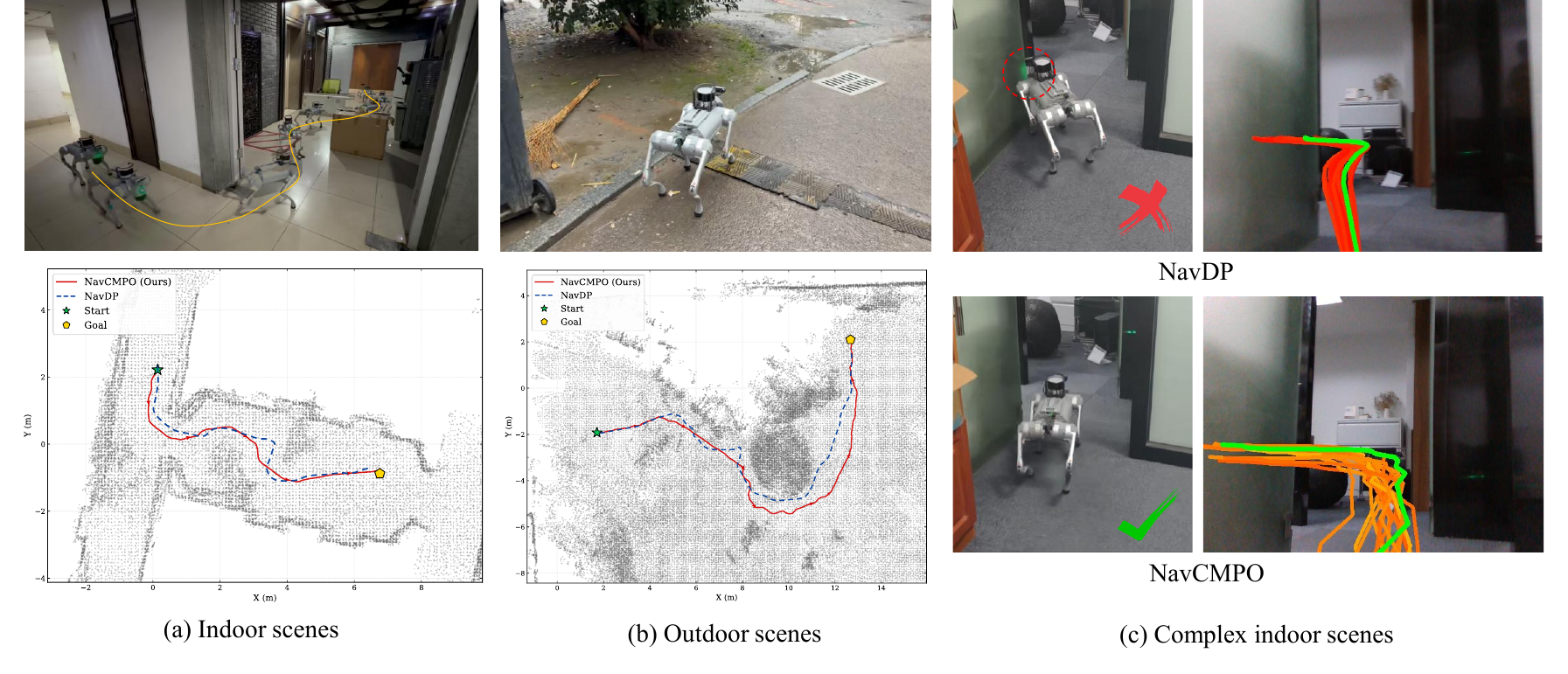}
\caption{Real-world Experiments. (a) Indoor scene: the top row shows the robot traversing a cluttered corridor; the bottom row shows the corresponding top-down trajectory on a point-cloud map, where the red solid line denotes NavCMPO (Ours) and the blue dashed line denotes NavDP. (b) Outdoor navigation: the robot navigates along an unstructured outdoor path with obstacles. (c) Complex indoor scene: NavDP collides with furniture and fails ($\times$), while NavCMPO successfully avoids the obstacle and reaches the goal (\checkmark).}
\label{fig:realworld}
\vspace{-0.5cm}
\end{figure*}

In fine-tuning stage, we conducted ablation experiments on the number of fine-tuning steps in reinforcement learning. Fig.~\ref{fig:reward} shows the training reward curves under different denoising steps. We take $\beta_p=5,\beta_s=50,\beta_g=20$ in eq.\ref{eq:reward}. The 5-step configuration achieves the highest and most stable reward growth, while the 1-step configuration stagnates, confirming that sufficient denoising steps are essential for effective PPO fine-tuning of MeanFlow policies.

\subsection{Real World Results}

We conducted real-world experiments using the Unitree Go2 quadruped robot, navigation strategies are generated with a D435i depth camera. Before real-world deployment, the policy was trained by reinforcement learning in a simulated environment using the Go2 model, followed by Sim-to-Real transfer. To evaluate the performance of our method, we designed point-goal navigation experiments across various scenarios, including simple indoor, complex indoor, and outdoor environments. In indoor and outdoor scenarios, we chose relatively open spaces with clearly defined obstacles. In complex indoor scenarios, we guided the robot to navigate through narrow spaces. The baseline algorithms selected for comparison include ViPlanner, NavDP, NavCMP, and NavCMPO.

\begin{table}[h]
\centering
\caption{Real World Experiments.}
\label{tab:real}
\begin{tabular}{lcccc}
\toprule
\multirow{2}{*}{Method} & Easy & Outdoor  & Hard & Average \\
 & SR  & SR  & SR  & SR    \\
\midrule
Viplanner & 4/10 &  5/10 &  1/10 & 33.3\% \\
NavDP & 7/10 &  7/10 &  3/10 & 56.7\% \\
NavCMP & 7/10 & 6/10 &  3/10 & 53.3\% \\
\textbf{NavCMPO} & \textbf{7/10} & \textbf{8/10} & \textbf{5/10} & 66.7\% \\
\bottomrule
\end{tabular}
\end{table}

As summarized in Table~\ref{tab:real}, ViPlanner achieves the lowest overall success rate (33.3\%), particularly in Hard scenes (1/10). NavDP and NavCMP perform comparably in Easy (7/10) and Hard (3/10) scenes, indicating that MeanFlow-based trajectory generation  achieves similar sim-to-real transfer capability as the DDPM-based approach. However, NavCMPO significantly outperforms all baselines in both Outdoor (8/10) and Hard (5/10) scenarios, achieving the highest average success rate of 66.7\%. As illustrated in Fig.~\ref{fig:realworld}(c), in complex indoor scenes where NavDP collides with obstacles and fails, NavCMPO successfully navigates through narrow passages. The trajectory visualization in Fig.~\ref{fig:realworld}(c) further reveals that NavCMPO generates safer trajectories with better obstacle clearance compared to NavDP, which tends to produce erratic paths near obstacles. The improvements in Hard and Outdoor scenes demonstrate that the combination of CGTR and PPO fine-tuning enhances the policy's robustness to real-world distributional shifts, particularly in scenarios requiring precise obstacle avoidance that were not adequately covered by the expert demonstrations.

\section{Conclusion}

In this paper, we presented NavCMPO, a two-stage framework for low-latency and adaptive visual navigation. By combining few-step MeanFlow generation with obstacle-aware representation learning and critic-guided trajectory refinement, NavCMPO maintains reliable obstacle avoidance while reducing inference latency. PPO fine-tuning further improves task performance, while critic adaptation and behavior-cloning regularization support stable specialization to a target embodiment. Under a matched training budget, NavCMPO improves the average success rate over the retrained NavDP baseline and transfers effectively to a Unitree Go2. The current evaluation is limited to a matched-compute setting and a single real-world robot embodiment, and it does not directly compare against applying the same RL fine-tuning procedure to NavDP. In addition, policy adaptation is currently performed in simulation because unrestricted online reinforcement learning on physical robots introduces safety risks. Future work will investigate broader cross-embodiment evaluation, explicit geometric-guidance baselines, and safe real-world policy adaptation.

\bibliographystyle{IEEEtran}
\bibliography{refer}

\end{document}